\pdfoutput=1

\documentclass[11pt]{article}

\usepackage[]{EMNLP2023}

\usepackage{times}
\usepackage{latexsym}

\usepackage{xcolor}         
\usepackage{colortbl}         
\usepackage{amssymb} 
\usepackage{hyperref}

\usepackage[T1]{fontenc}

\usepackage[utf8]{inputenc}

\usepackage{microtype}

\usepackage{inconsolata}

\usepackage{graphicx}
\usepackage{amsmath}
\usepackage{nccmath}
\usepackage{amssymb}
\usepackage{mathtools}
\usepackage{multirow}
\usepackage{xcolor}

\title{90\% F1 Score in Relational Triple Extraction: Is it Real ?}

\author{Pratik Saini \and Samiran Pal \and Tapas Nayak \and Indrajit Bhattacharya  \\
        TCS Research, India \\
        \texttt{\{pratik.saini,samiran.pal,nayak.tapas,b.indrajit\}}@tcs.com\\}

\begin{document}
\maketitle

\begin{abstract}
Extracting relational triples from text is a crucial task for constructing knowledge bases. Recent advancements in joint entity and relation extraction models have demonstrated remarkable F1 scores ($\ge 90\%$) in accurately extracting relational triples from free text. However, these models have been evaluated under restrictive experimental settings and unrealistic datasets. They overlook sentences with zero triples (zero-cardinality), thereby simplifying the task. In this paper, we present a benchmark study of state-of-the-art joint entity and relation extraction models under a more realistic setting. We include sentences that lack any triples in our experiments, providing a comprehensive evaluation. Our findings reveal a significant decline (approximately 10-15\% in one dataset and 6-14\% in another dataset) in the models' F1 scores within this realistic experimental setup. Furthermore, we propose a two-step modeling approach that utilizes a simple BERT-based classifier. This approach leads to overall performance improvement in these models within the realistic experimental setting.
\end{abstract}

\section{Introduction}

A crucial aspect of the relation extraction task involves the identification of sentences that lack any relational triples. This aspect naturally arises in real-world relation extraction scenarios. For instance, when extracting knowledge graph triples from online text, the majority of sentences may not mention any such triples. Although this aspect has been explored in other NLP tasks, such as machine reading comprehension, where models should correctly identify when a given passage lacks an answer rather than providing an incorrect one~\cite{Rajpurkar2018KnowWY, Kundu2018ANA, Sulem2021DoWK}, it has not received sufficient attention in recent relation extraction research.

There are two distinct approaches for entity and relation extraction: Classification approach and joint approach. In the classification approach \cite{hoffmann2011knowledge, zeng2014relation, zeng2015distant, nayak2019effective, jat2018attention}, entities are already given and models focus on classifying the relations among pairs of entities. This approach includes sentences with zero triples in the experiments, where the relation among all entity pairs in such sentences is labeled as a `None' relation. On the other hand, the joint extraction approach \cite{zeng2018copyre, hrlre2019takanobu, nayak2020ptrnetdecoding, Wei2020ANC, Wang2020TPLinkerSJ, Zheng2021PRGCPR, Li2021TDEERAE, Wei2020ANC, Yan2021APF, Shang2022OneRelJE} involves models extracting both entities and relations simultaneously. However, in this approach, sentences with zero triples are not considered in the experiments, which makes the task significantly easier. Consequently, recent joint extraction models achieve exceptionally high F1 scores on benchmark datasets.


\begin{table*}[pt]
\small
\centering
\begin{tabular}{|l|l|}
\hline
\multicolumn{1}{|c|}{Sentence}                                                                                                                                                                                                                                            & \multicolumn{1}{c|}{Triples}                                                                                                                      \\ \hline
\begin{tabular}[c]{@{}l@{}}Paul Allen , a \textcolor{pink}{co-founder} of Microsoft , paid the bills for aircraft \\ designer Burt Rutan to develop SpaceShipOne , the craft that \\ won the \$ 10 million Ansari X Prize last year for reaching \\ suborbital space .\end{tabular}         & \begin{tabular}[c]{@{}l@{}}Microsoft ; Paul Allen ; /business/company/founders \\ Paul Allen ; Microsoft ; /business/person/company\end{tabular} \\ \hline
\begin{tabular}[c]{@{}l@{}}But Schaap seems as comfortable in that role as Joe Buck , the \\ Fox baseball and football sportscaster who so clearly benefited \\ from learning beside his \textcolor{pink}{father} , Jack Buck , the late voice of \\ the St. Louis Cardinals .\end{tabular} & Jack Buck ; Joe Buck ; /people/person/children                                                                                                   \\ \hline
\end{tabular}
\caption{Examples of relation clue tokens (in \textcolor{pink}{Pink}) for determining the presence of a relation in the sentences.}
\label{tab:trigger_token}
\end{table*}

In this study, our objective is to assess the performance of state-of-the-art relational triples extraction models when sentences with zero triples are included. To achieve this, we conduct comprehensive experiments using the widely used New York Times (NYT) datasets. We evaluate a total of 9 recent state-of-the-art models in an end-to-end fashion. The results of our experiments reveal a significant decline in the performance of these models under this 
experimental setting. Across all of the evaluated models, we observe an approximate drop of 10-15\% in the F1 score in one dataset, and a drop of around 6-14\% in another dataset. These findings highlight the challenges posed by sentences without triples and emphasize the need for improved approaches to handle such cases effectively.

Additionally, we have identified that sentences often contain clue tokens that can be leveraged to detect the presence of relations, even without identifying the corresponding entities. We include such examples in Table \ref{tab:trigger_token} for illustrations. Building upon this observation, we introduce a BERT-based zero-cardinality classifier (ZCC) model that effectively filters out sentences with zero triples. We explore both binary classification and multi-class multi-label (MCML) classification approaches for this purpose. To tackle the task at hand, we propose a two-step modeling approach. In the first step, we employ the ZCC model to classify the sentences, determining whether they contain zero triples or not. In the second step, we utilize the outputs of the ZCC model to guide the 9 state-of-the-art triples extraction models, effectively solving the task. Notably, our experimental results demonstrate that this two-step approach outperforms or achieves competitive performance compared to end-to-end modeling in this novel setting of the task. Furthermore, it offers advantages in terms of training time for the models\footnote{Any code or data related to this paper will be made available at \url{https://github.com/pratiksaini4/ZeroCardinalityImpactOnRE}.}.


\section{End-to-End Modeling of Relation Extraction with Zero-Cardinality}

For our experiments, we select nine state-of-the-art joint entity and relation extraction models: PtrNet \cite{nayak2020ptrnetdecoding}, TPLinker \cite{Wang2020TPLinkerSJ}, CasRel \cite{Wei2020ANC}, TDEER \cite{Li2021TDEERAE}, PRGC \cite{Zheng2021PRGCPR}, PFN \cite{Yan2021APF}, GRTE \cite{Ren2021ANG}, OneRel \cite{Shang2022OneRelJE}, and BiRTE \cite{Ren2022ASB}. All of these models utilize BERT \cite{Devlin2019BERTPO} as an encoder. For our experiments with the NYT24* dataset, where sentences are cased, we utilize the BERT\_base\_cased model. On the other hand, for the NYT29* dataset, where sentences are uncased, we use the BERT\_base\_uncased model.

PtrNet \cite{nayak2020ptrnetdecoding} adopts a sequence-to-sequence (seq2seq) approach, extracting triples uniformly regardless of the relations involved. The remaining models employ relation-specific sequence or matrix labeling methods to extract triples. Originally, these models are trained solely on sentences containing one or more triples, excluding sentences with zero triples from their training and test datasets. However, we adapt these models to handle sentences with zero triples as well. In the case of sequence labeling or matrix labeling approaches, all tokens in the zero-cardinality sentences are labeled with the 'O' tag (representing the "other" tag). For sequence generation approaches (such as seq2seq), the decoder generates the "end of sequence" (EOS) tag as the first token, indicating the absence of any relational triple in the sentence.

Below is a brief description of each of these models. We employ the same hyper-parameters as specified in their respective papers.

\subsection{PtrNet \cite{nayak2020ptrnetdecoding}}

This model utilizes a seq2seq framework with pointer network-based decoding for joint entity and relation extraction. Each triple is represented by the start and end indices of the subject and object entities in the sentence, along with the corresponding relation class label. To generate the complete triple, their decoding framework extracts four indexes at each time step, capturing the subject and object entities as well as the relation between them. This enables the model to incrementally construct the entire triple. For a fair comparison with other state-of-the-art (SOTA) models, the original BiLSTM encoder is replaced with BERT, a powerful language representation model. This integration of BERT into the model ensures compatibility and consistency with the advancements in the field, allowing for more accurate and robust results.

\subsection{CasRel \cite{Wei2020ANC}}

CasRel employs a two-stage extraction process for relation extraction. In the first stage, it utilizes a 0/1 tagging scheme to identify all subject entities present in the text. This initial stage focuses on accurately identifying and labeling the subject entities involved in the relations. In the subsequent stage, for each subject entity and for each relation, CasRel applies another round of 0/1 tagging to identify the corresponding object entities. This object tagging process is iterative and carried out sequentially for each subject entity. By performing this iterative tagging approach, CasRel ensures comprehensive identification of the object entities associated with each subject entity, enabling a more precise extraction of relational triples.

\subsection{TPLinker \cite{Wang2020TPLinkerSJ}}

TPLinker also adopts a sequence labeling approach for the relation extraction task. However, to effectively address the challenges posed by overlapping triples, it employs a separate sequence labeling process for each relation. To link the tokens within the sentence, TPLinker utilizes a handshaking tagging scheme. It constructs a matrix representing the tokens in the sentence, where the rows and columns correspond to the tokens. The handshaking tags are employed to establish connections between tokens. Initially, TPLinker identifies all entities in the sentence using the `EH-ET' (entity head to entity tail) tag. In the matrix, a cell with a value of 1 indicates that the token in the corresponding row represents the start of an entity, while the token in the column represents the end of the entity. Additionally, TPLinker employs two other handshaking tags, namely SH-OH' (subject head to object head) and ST-OT' (subject tail to object tail). These tags are used to link the subject and object entities for each specific relation. Separate matrices are tagged for each relation using these handshaking tags. By applying this approach, TPLinker effectively links the subject and object entities for each relation, enabling accurate extraction of relational triples. The initial set of entities obtained from the `EH-ET' tagging stage serves to filter out unwanted triples extracted during the relation-specific tagging stage.

\subsection{TDEER \cite{Li2021TDEERAE}}

This task employs a multi-stage sequence labeling approach. In the initial stage, a 0/1 tagging scheme is utilized to extract subject and object entities. Additionally, a multi-label classification technique is employed to identify all possible relations present in the sentence. In the subsequent stage, for each subject entity and relation pair, the start position of the corresponding object entity is identified. If this start position aligns with any of the object entities extracted in the first stage, the triple is considered valid and retained. Conversely, if no match is found, the triple is deemed invalid and discarded. This rigorous validation process ensures the accuracy and reliability of the extracted triples.

\subsection{PRGC \cite{Zheng2021PRGCPR}}

In this model, the first step involves identifying a set of potential relations within the sentence, as well as establishing a global correspondence matrix between the subject and object entities. In the subsequent stage, relation-specific sequence taggers are employed to label the subject and object entities accordingly. These taggers provide fine-grained annotations, enabling precise identification of the entities involved. Finally, the global correspondence matrix is utilized to make informed decisions regarding which triples to accept or discard. By considering the interplay between the subject and object entities and their respective relations, the model ensures the selection of valid and meaningful triples while discarding any irrelevant or incorrect ones.

\subsection{GRTE \cite{Ren2021ANG}}

This approach utilizes a table filling method where separate tables are maintained for each relation. Each cell in the table represents whether a token pair is associated with the corresponding relation or not. These tables are populated using local features or the historical information of a limited number of token pairs. GRTE enhances the table-filling by incorporating two types of global features. The first type pertains to the global association of entity pairs, while the second type focuses on relations. GRTE initially generates a table feature for each relation. Subsequently, these table features for all relations are combined, resulting in the creation of a subject-related global feature and an object-related global feature. These global features are then utilized iteratively to refine the individual table features. By employing this refined table-filling approach, all triples can be extracted based on the information stored in the populated tables. This method enables the accurate and comprehensive extraction of relational triples.

\subsection{PFN \cite{Yan2021APF}}

The model consists of two main modules: the Named Entity Recognition (NER) module and the Relation Extraction (RE) module. In the NER module, all named entities in the sentences are extracted, capturing their complete spans. This module focuses on identifying and delineating entities present in the text. The RE module operates separately for each relation. It employs matrix labeling techniques to identify the starting tokens of subject and object entity pairs. The full span of these entities is obtained from the entities previously identified by the NER module. By leveraging the information provided by the NER module, the RE module can accurately determine the boundaries and positions of the subject and object entities for each relation.

\subsection{OneRel \cite{Shang2022OneRelJE}}

The approach utilized in this task is a relation-specific horns-tagging method. For each relation in the set of relations, a matrix is maintained, consisting of four types of tags: `HB-TB', `HB-TE', `HE-TE', and `O'. Here, `H/T' represents the head or tail entity, while `B/E' denotes the beginning and ending of an entity. The rows of this matrix correspond to the head entity tokens, while the columns correspond to the tail entity tokens derived from the source text. Following the tagging of these matrices, a scoring-based classifier is employed to iterate through all possible combinations and discard triples with low confidence scores. This process enables the identification and retention of high-quality triples based on their associated confidence scores.

\subsection{BiRTE \cite{Ren2022ASB}}

This model employs a multi-stage bidirectional tagging-based mechanism. In the initial stage, the model focuses on identifying subject and object entities. Subsequently, in the second stage, it further refines the identification of object entities based on the previously identified subject entities, and vice versa. Finally, in the last stage, subject-object pairs are classified based on their respective relations. All these stages are trained together as a single model, ensuring a comprehensive and integrated approach to relation extraction.

\section{Two-step Modeling of Relation Extraction with Zero-Cardinality} 

We have observed that most relational triples in sentences are associated with specific clue tokens. While this may not always hold true due to the distant supervision used in creating the NYT datasets, it is applicable to many cases. We have included relevant examples in Table \ref{tab:trigger_token}. Based on this observation, we aim to investigate whether a BERT-based classification model can learn to identify the presence of relational triples in these sentences using the clue tokens, without requiring knowledge of the specific entities involved in the triples.

To accomplish this, we feed the sentences with a `CLS' prefix token (CLS $w_1$ $w_2$ ..... $w_n$) into a pre-trained BERT\_{base} model with a hidden dimension of $h$. We utilize the vector representation of the `CLS' token to determine whether the sentence contains any relational triples or not. We refer to this classifier as the zero-cardinality classifier (ZCC).

We explore two distinct approaches for this classifier:

(i) The first approach involves binary classification to determine whether a sentence contains any triples or not. However, in this approach, we do not explicitly utilize the set of relations.

(ii) The second approach employs a multi-class multi-label (MCML) classification, which focuses on identifying the specific relations within the relation set. Sentences without any triples are assigned no positive labels.

\begin{table*}[ht]
\small
\centering
\begin{tabular}{l|ccc|ccc}
\hline
                                                                                 & \multicolumn{3}{c|}{NYT24*} & \multicolumn{3}{c}{NYT29*} \\ 
\multicolumn{1}{c|}{}                                                           & Train    & Validation    & Test   & Train    & Validation    & Test   \\ \hline

\begin{tabular}[c]{@{}l@{}}\#sentences with $>$=1 triples\end{tabular}    & 56,196   & 5,000  & 5,000  & 63,306   & 7,033  & 4,006 \\ 
\begin{tabular}[c]{@{}l@{}}\#triples in above sentences\end{tabular} & 88,366  & 8,489  & 8,120 & 78,973  & 8,766  & 5,859   \\ 
 
\begin{tabular}[c]{@{}l@{}}\#sentences with zero triples\end{tabular} & 145,767  & 4,969  & 4,969 & 177,861  & 4,940  & 4,601  \\ \hline
\end{tabular}
\caption{The statistics of the NYT24* and NYT29* datasets.}
\label{tab:data_stat}
\end{table*}

To begin, we train the classifier on the `WZ' training dataset, while training the joint extraction models on the `NZ' training set. During the inference phase, if the classifier model indicates the presence of triples in a test instance, we subsequently pass it to the joint extraction models to extract the exact triples. This two-step process enables us to effectively filter out sentences that do not contain any triples.

\begin{figure}[t]
    \centering
    \includegraphics[scale=0.5]{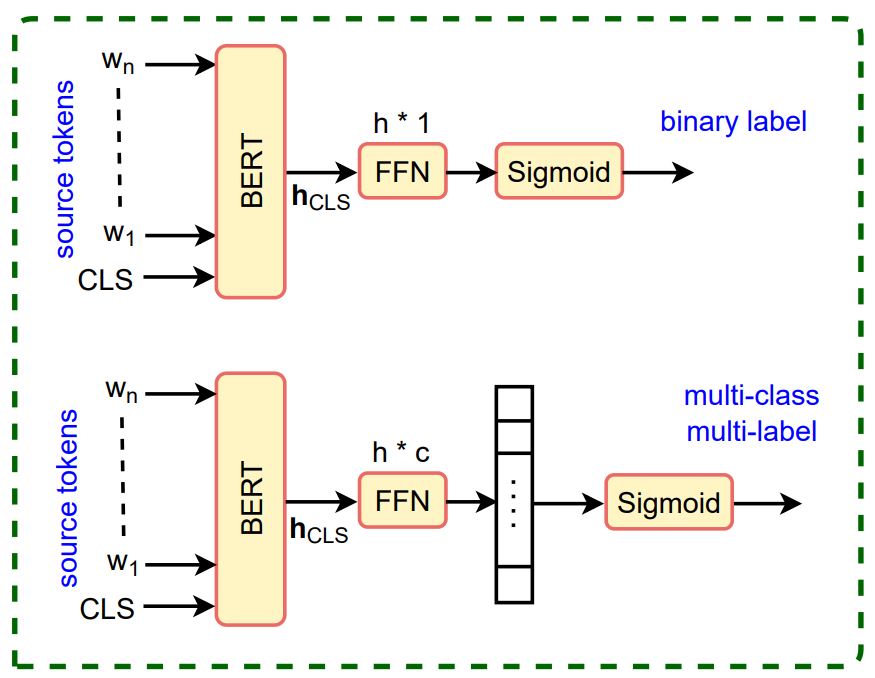}
    \caption{Architecture of our zero-cardinality classifier. $c$ is the number of relations. }
    \label{fig:zero-card-classifier}
\end{figure}

We include the architecture of our proposed zero-cardinality classifier in Fig \ref{fig:zero-card-classifier}. We use binary cross-entropy loss and AdamW \cite{adamw2019ilya} optimizer to update the model parameters. We use mini-batch size of 16 and an early stop criterion during training. Our experiments have demonstrated that this two-step approach significantly enhances the overall performance of the joint models on the test set, encompassing sentences both with and without triples.

\section{Datasets Preparation \& Evaluation Metric}

The New York Times (NYT) dataset holds significant importance as a benchmark for relation extraction. Several studies \cite{zeng2018copyre, hrlre2019takanobu, nayak2020ptrnetdecoding} utilize the derived NYT29 and NYT24 datasets, which originate from the original NYT10 \cite{riedel2010modeling} and NYT11 \cite{hoffmann2011knowledge} training corpus, respectively. \citet{zeng2018copyre, hrlre2019takanobu, nayak2020ptrnetdecoding} exclude sentences without triples and partition the dataset into training, validation, and test sets. Subsequent research papers \cite{Wei2020ANC, Wang2020TPLinkerSJ, Zheng2021PRGCPR, Li2021TDEERAE, Wei2020ANC, Yan2021APF, Shang2022OneRelJE} build upon this modified version of the datasets, which is comparatively easier, and achieve exceptionally high F1 scores on these datasets. This trend reflects the prevalence of simplified datasets in recent works, potentially overestimating the performance of relation extraction models when faced with more realistic scenarios.

In order to enhance the realism of the joint extraction task, we augment the NYT29 and NYT24 datasets by incorporating sentences with zero triples from the original NYT10 and NYT11 training corpus, respectively. These augmented datasets are referred to as NYT29* and NYT24* hereafter. The specifics regarding the training, validation, and test splits of the NYT24* and NYT29* datasets can be found in Table \ref{tab:data_stat}.

To evaluate the state-of-the-art (SOTA) models, we conduct experiments using two distinct training and test settings. These settings are as follows:

(i) \textbf{NoZero (NZ)}: In this setting, only sentences containing one or more triples are included for training and testing purposes.

(ii) \textbf{WithZero (WZ)}: This setting encompasses the sentences from the NZ set, along with additional sentences with zero triples from the corresponding original NYT datasets.

By employing these two different experimental designs, we aim to gain insights into the robustness of the joint extraction models and their ability to handle different scenarios.

\subsection{Evaluation Metric}

For evaluating the performance of the state-of-the-art (SOTA) models, we employ triple-level precision, recall, and F1 score as the evaluation metrics. In order to determine the correctness of an extracted triple, we compare it with the ground truth triple. A triple is considered correct if both the corresponding entities and the relation match accurately. In the case of an 'Exact' match, we require the full span of the entities to match precisely, as specified in the respective papers. However, in the case of a 'Partial' match, we only compare the first or last token of the entities with the ground truth.

\section{Results \& Discussion}

\begin{table*}[ht]
\small
\centering
\begin{tabular}{c|c|l|lll|lll|ll}
\hline
\multicolumn{1}{l|}{}     & \multicolumn{1}{l|}{}                                                             & Test setting → & \multicolumn{3}{c|}{NZ}                      & \multicolumn{3}{c|}{WZ}                      &            &               \\ \hline
\multicolumn{1}{l|}{}     & \multicolumn{1}{l|}{\begin{tabular}[c]{@{}l@{}}Training\\ setting ↓\end{tabular}} & Model          & Prec. & Rec.  & F1                           & Prec. & Rec.  & F1                           & \% point ↓ &               \\ \hline
                          &                                                                                   & OneRel         & 0.926 & 0.918 & {\color[HTML]{009901} 0.922} & 0.678 & 0.918 & 0.780                        & 14.2       &               \\
                          &                                                                                   & BiRTE          & 0.914 & 0.920 & {\color[HTML]{009901} 0.917} & 0.628 & 0.920 & 0.747                        & 17.0       &               \\
                          &                                                                                   & TDEER          & 0.922 & 0.908 & {\color[HTML]{009901} 0.915} & 0.644 & 0.908 & 0.754                        & 16.1       &               \\
                          &                                                                                   & PRGC           & 0.918 & 0.884 & {\color[HTML]{009901} 0.901} & 0.670 & 0.884 & 0.762                        & 13.9       &               \\
                          &                                                                                   & GRTE           & 0.929 & 0.924 & {\color[HTML]{009901} 0.926} & 0.645 & 0.924 & 0.760                        & 16.6       &               \\
                          &                                                                                   & PtrNet         & 0.898 & 0.894 & {\color[HTML]{009901} 0.896} & 0.538 & 0.894 & 0.671                        & 22.5       &               \\
                          &                                                                                   & CasRel         & 0.894 & 0.890 & {\color[HTML]{009901} 0.892} & 0.612 & 0.890 & 0.725                        & 16.7       &               \\
                          &                                                                                   & TPLinker       & 0.913 & 0.917 & {\color[HTML]{009901} 0.915} & 0.643 & 0.917 & 0.756                        & 15.9       &               \\
                          & \multirow{-9}{*}{NZ}                                                              & PFN*           & 0.892 & 0.919 & {\color[HTML]{009901} 0.905} & 0.557 & 0.919 & 0.694                        & 21.1       &               \\ \cline{2-11} 
                          &                                                                                   & OneRel         & 0.926 & 0.773 & 0.843                        & 0.828 & 0.773 & {\color[HTML]{FE0000} 0.800} & 4.3        & \textbf{12.2} \\
                          &                                                                                   & BiRTE          & 0.898 & 0.858 & 0.878                        & 0.786 & 0.858 & {\color[HTML]{FE0000} 0.820} & 5.8        & \textbf{9.7}  \\
                          &                                                                                   & TDEER          & 0.914 & 0.905 & 0.909                        & 0.637 & 0.905 & {\color[HTML]{FE0000} 0.748} & 16.1       & \textbf{16.7} \\
                          &                                                                                   & PRGC           & 0.905 & 0.777 & 0.836                        & 0.791 & 0.777 & {\color[HTML]{FE0000} 0.784} & 5.2        & \textbf{11.7} \\
                          &                                                                                   & GRTE           & 0.920 & 0.769 & 0.838                        & 0.824 & 0.769 & {\color[HTML]{FE0000} 0.796} & 4.2        & \textbf{13.0} \\
                          &                                                                                   & PtrNet         & 0.932 & 0.697 & 0.798                        & 0.838 & 0.697 & {\color[HTML]{FE0000} 0.761} & 3.7        & \textbf{13.5} \\
                          &                                                                                   & CasRel         & 0.915 & 0.878 & 0.896                        & 0.643 & 0.878 & {\color[HTML]{FE0000} 0.742} & 15.4       & \textbf{15.0} \\
                          &                                                                                   & TPLinker       & 0.923 & 0.808 & 0.861                        & 0.823 & 0.807 & {\color[HTML]{FE0000} 0.815} & 4.6        & \textbf{10.0} \\
\multirow{-18}{*}{NYT24*} & \multirow{-9}{*}{WZ}                                                              & PFN*           & 0.910 & 0.732 & 0.812                        & 0.804 & 0.732 & {\color[HTML]{FE0000} 0.766} & 4.6        & \textbf{13.9} \\ \hline
                          &                                                                                   & OneRel         & 0.805 & 0.726 & {\color[HTML]{009901} 0.763} & 0.528 & 0.726 & 0.611                        & 15.2       &               \\
                          &                                                                                   & BiRTE          & 0.794 & 0.724 & {\color[HTML]{009901} 0.757} & 0.484 & 0.724 & 0.580                        & 17.7       &               \\
                          &                                                                                   & TDEER          & 0.813 & 0.707 & {\color[HTML]{009901} 0.756} & 0.530 & 0.707 & 0.606                        & 15.0       &               \\
                          &                                                                                   & PRGC           & 0.807 & 0.701 & {\color[HTML]{009901} 0.750} & 0.509 & 0.701 & 0.590                        & 16.0       &               \\
                          &                                                                                   & GRTE           & 0.804 & 0.726 & {\color[HTML]{009901} 0.763} & 0.492 & 0.726 & 0.587                        & 17.6       &               \\
                          &                                                                                   & PtrNet         & 0.790 & 0.710 & {\color[HTML]{009901} 0.748} & 0.394 & 0.710 & 0.507                        & 24.1       &               \\
                          &                                                                                   & CasRel         & 0.795 & 0.712 & {\color[HTML]{009901} 0.751} & 0.488 & 0.712 & 0.579                        & 17.2       &               \\
                          &                                                                                   & TPLinker       & 0.805 & 0.718 & {\color[HTML]{009901} 0.759} & 0.456 & 0.718 & 0.558                        & 20.1       &               \\
                          & \multirow{-9}{*}{NZ}                                                              & PFN*           & 0.777 & 0.720 & {\color[HTML]{009901} 0.748} & 0.474 & 0.720 & 0.572                        & 17.6       &               \\ \cline{2-11} 
                          &                                                                                   & OneRel         & 0.841 & 0.657 & 0.738                        & 0.755 & 0.657 & {\color[HTML]{FE0000} 0.703} & 3.5        & \textbf{6.0}  \\
                          &                                                                                   & BiRTE          & 0.833 & 0.663 & 0.738                        & 0.698 & 0.663 & {\color[HTML]{FE0000} 0.680} & 5.8        & \textbf{7.7}  \\
                          &                                                                                   & TDEER          & 0.788 & 0.708 & 0.746                        & 0.536 & 0.708 & {\color[HTML]{FE0000} 0.611} & 13.5       & \textbf{14.5} \\
                          &                                                                                   & PRGC           & 0.842 & 0.639 & 0.727                        & 0.755 & 0.639 & {\color[HTML]{FE0000} 0.692} & 3.5        & \textbf{5.8}  \\
                          &                                                                                   & GRTE           & 0.840 & 0.624 & 0.716                        & 0.759 & 0.623 & {\color[HTML]{FE0000} 0.684} & 3.2        & \textbf{7.9}  \\
                          &                                                                                   & PtrNet         & 0.876 & 0.620 & 0.726                        & 0.720 & 0.620 & {\color[HTML]{FE0000} 0.666} & 6.0        & \textbf{8.2}  \\
                          &                                                                                   & CasRel         & 0.807 & 0.708 & 0.754                        & 0.541 & 0.708 & {\color[HTML]{FE0000} 0.613} & 14.1       & \textbf{13.8} \\
                          &                                                                                   & TPLinker       & 0.775 & 0.636 & 0.698                        & 0.686 & 0.636 & {\color[HTML]{FE0000} 0.660} & 3.8        & \textbf{9.9}  \\
\multirow{-18}{*}{NYT29*} & \multirow{-9}{*}{WZ}                                                              & PFN*           & 0.833 & 0.600 & 0.697                        & 0.748 & 0.600 & {\color[HTML]{FE0000} 0.666} & 3.1        & \textbf{8.2}  \\ \hline
\end{tabular}
\caption{Performance of the joint extraction models in the end-to-end approach on the NYT24* and NYT29* datasets with different train/test settings. * marked models are evaluated using partial entity matching as per their paper. F1 score in {\color[HTML]{009901} green} color are the results obtained without zero-cardinality sentences. F1 score in \textcolor{red}{red} color are the results obtained with zero-cardinality sentences. The \% point $\downarrow$ numbers in \textbf{bold} are the difference between the F1 scores in {\color[HTML]{009901} green} and \textcolor{red}{red}.}
\label{tab:results}
\end{table*}

\begin{table*}[ht]
\centering
\begin{tabular}{l|lll|lll}
\hline
                        & \multicolumn{3}{c|}{NYT24*} & \multicolumn{3}{c}{NYT29*} \\ \hline
\textbf{}               & Prec.   & Rec.    & F1      & Prec.   & Rec.    & F1     \\ \hline
ZCC$_{binary}$                  & 0.887   & 0.867   & 0.877   & 0.801   & 0.888   & 0.842  \\
ZCC$_{MCML}$ & 0.881   & 0.884   & 0.883   & 0.823   & 0.824   & 0.823  \\ \hline
\end{tabular}
\caption{Performance of the zero cardinality classifier (ZCC) model on NYT24* and NYT29* datasets in the binary classification and multi-class multi-label classification (MCML) settings.}
\label{tab:classification-results-table}
\end{table*}

\begin{table*}[ht]
\centering
\begin{tabular}{l|llll|llll}
\hline
         & \multicolumn{4}{c|}{NYT24*}                  & \multicolumn{4}{c}{NYT29*}    \\ \hline
         & \multicolumn{4}{c|}{multi-class multi-label} & \multicolumn{4}{c}{binary}    \\ \hline
Model    & Prec.     & Rec.      & F1        & \% ↑     & Prec. & Rec.  & F1    & \% ↑  \\ \hline
OneRel   & 0.832     & 0.836     & 0.834     & 3.43     & 0.740 & 0.664 & 0.700 & -0.27 \\
BiRTE    & 0.819     & 0.839     & 0.829     & 0.85     & 0.679 & 0.663 & 0.671 & -0.95 \\
TDEER    & 0.830     & 0.830     & 0.830     & 8.23     & 0.749 & 0.649 & 0.696 & 8.52  \\
PRGC     & 0.822     & 0.811     & 0.816     & 3.26     & 0.744 & 0.645 & 0.691 & -0.14 \\
GRTE     & 0.835     & 0.842     & 0.839     & 4.30     & 0.740 & 0.661 & 0.699 & 1.41  \\
PtrNet   & 0.806     & 0.815     & 0.811     & 4.95     & 0.677 & 0.650 & 0.663 & -0.33 \\
CasRel   & 0.807     & 0.812     & 0.810     & 6.73     & 0.676 & 0.653 & 0.665 & 5.13  \\
TPLinker & 0.816     & 0.839     & 0.828     & 1.23     & 0.681 & 0.656 & 0.668 & 0.81  \\
PFN*     & 0.805     & 0.833     & 0.818     & 5.20     & 0.726 & 0.658 & 0.690 & 2.42  \\ \hline
\end{tabular}
\caption{Performance of the SOTA models in the two-step modeling on the relational triple extraction task with zero-cardinalty sentences. At the first-step, we use multi-class multi-label classification for NYT24* dataset and binary classification for NYT29* dataset.}
\label{tab:classification-pipeline-end-task-performance}
\end{table*}

To begin our analysis, we assess the performance of state-of-the-art (SOTA) end-to-end models under the new experiment settings, which now include sentences with zero cardinality. The results of these experiments are presented in Table \ref{tab:results}. Initially, we train these models solely on the `NZ' sentences and evaluate their performance on both the `NZ' and `WZ' sentences. Upon evaluation, we observe a significant decline in the F1 score on the WZ' sentences compared to the NZ' sentences. Across the NYT24* and NYT29* datasets, the F1 score experiences a decrease of approximately 14-24\%. Furthermore, the precision score for all these models exhibits a sharp drop, as they extract triples from sentences that do not contain any triples. This outcome is expected since the models have not been exposed to any examples featuring zero triples during the training phase.

Next, we proceed to train these models using the `WZ' sentences. Upon analysis, we note that their performance on the `NZ' sentences experiences a decline of 4-8\%, with the exception of the TDEER and CasRel models. Interestingly, the TDEER and CasRel models exhibit comparable performance on the `NZ' test set, regardless of whether they were trained on `NZ' or `WZ' training data. However, the introduction of sentences with zero triples during the training process tends to confuse these models, leading to a negative impact on their recall. Consequently, the models struggle to accurately extract valid triples due to the presence of such adversarial examples. Furthermore, in this training setting, we observe an improvement of 2-8\% in the models' performance on `WZ' sentences. Nevertheless, the best F1 score reported on the stringent `NZ' test set for NYT24* is 0.926 (achieved by the GRTE model). In contrast, the best F1 score attained on the `WZ' test set for NYT24* is 0.82 (achieved by the BiRTE model). This signifies a 10\% drop in the best F1 score when transitioning to the experiment's more diverse setting. Similarly, we observe a 6\% decrease in the best achieved F1 scores on the `WZ' test set for NYT29* compared to the `NZ' test set.

Next, we delve into the analysis of the impact of our proposed two-step approach for this task. The first step involves utilizing the zero-cardinality classifier to predict sentences with zero cardinality, i.e., sentences that either contain triples or do not. The performance of the classification model using both binary and multi-class multi-label (MCML) classification is provided in Table \ref{tab:classification-results-table}. The classification model was trained on `WZ' sentences for both the NYT24* and NYT29* datasets. Both binary classification and multi-class multi-label classification demonstrate competitive performance on both datasets. Multi-class multi-label classification exhibits slightly higher performance on the NYT24* dataset, while binary classification yields marginally better results on the NYT29* dataset.

In the second step of our two-step approach, only the sentences predicted by the classification model to have existing triples are passed on to the triple extraction model. For this step, we train the state-of-the-art (SOTA) models exclusively on the `NZ' sentences to facilitate triple extraction. In Table \ref{tab:classification-pipeline-end-task-performance}, we present the comprehensive performance evaluation of the state-of-the-art (SOTA) model using the two-step approach for the triple extraction task. For the NYT24* dataset, we utilize the multi-class multi-label classifier, while for the NYT29* dataset, we employ the binary classification approach for zero-cardinality prediction.

Our observations reveal an improvement of approximately $\sim 8\%$ in the `WZ' sentences for both the NYT24* and NYT29* datasets when employing the two-step approach compared to the end-to-end approach. Specifically, for the NYT24* dataset, all SOTA models exhibit enhanced performance with the two-step approach over the end-to-end approach. However, for the NYT29* dataset, the performance is not consistently improved. In the case of four models (OneRel, BiRTE, PRGC, and PtrNet), we observed a minor drop of up to $\sim 1\%$ with the two-step approach.

Overall, we conclude that the two-step approach either improves the performance of these models or achieves competitive performance when compared to the end-to-end approach in this new experimental setting for relation extraction.

\subsection{Training Time of the Models}

Table \ref{tab:avg-epoch-time} presents the training time of various models used in our experiments. All training was conducted on an NVIDIA A100 GPU with 20 GB GPU memory. Our two-step approach for the relation extraction task in this new setting offers advantages over the end-to-end approach.

\begin{table}[ht]
\small
\centering
\begin{tabular}{l|cc|cc}
\hline
       & \multicolumn{2}{c|}{NYT24*} & \multicolumn{2}{c}{NYT29*} \\ \hline
       & NZ           & WZ          & NZ          & WZ          \\ \hline
Onrel  & 18.33        & 68.35       & 21.05       & 83.06       \\
BiRTE  & 6.67         & 25.49       & 6.24        & 32.74       \\
TDEER  & 43.51        & 50.85       & 45.11       & 63.68       \\
PRGC   & 20.25        & 56.70       & 18.96       & 62.87       \\
GRTE   & 20.70        & 65.08       & 21.87       & 80.51       \\ 
PtrNet & 17.17        & 41.03       & 12.24       &  24.30           \\
CasRel & 18.23        & 65.20        & 20.54       & 75.95       \\
TPLinker & 26.19        & 122.65       & 43.40       & 168.91      \\
PFN*     & 22.40        & 317.18       & 188.68      & 393.35      \\ \hline
ZCC$_{binary}$ &   -           & 14.67       &   -    &  25.55           \\
ZCC$_{MCML}$   &   -           & 14.49       &   -    & 25.65           \\ \hline
\end{tabular}
\caption{Training time of the models. First 9 rows are avg. training epoch time (in minutes) of five SOTA models on the `NZ' and `WZ' training data. Last two rows are avg. training time of the zero cardinality classification (ZCC) models with WZ training data.}
\label{tab:avg-epoch-time}
\end{table}

The training time for the SOTA models solely using `NZ' data is considerable, primarily due to their utilization of BERT as the sentence encoder. However, when we incorporate sentences with zero triples in the training process (which account for almost three times the number of sentences with triples, as shown in Table \ref{tab:data_stat}), the training time significantly increases for all models (refer to Table \ref{tab:avg-epoch-time}).

On the contrary, the zero-cardinality classifier only needs to be trained once for all models, resulting in substantial time savings. Additionally, training the zero-cardinality classifier itself is relatively quick due to its simple architecture.

\section{Related Work}

Extracting relational triples from text is a crucial task for constructing new knowledge bases or enhancing existing ones. In their efforts to address this task, \citet{mintz2009distant,riedel2010modeling,hoffmann2011knowledge} employed feature-based classification models. More recently, \citet{zeng2014relation,zeng2015distant} utilized CNN models, which automatically extract features, for this purpose. \citet{huang2016attention,jat2018attention,nayak2019effective} incorporated attention mechanisms into their models to enhance performance. Approaches such as \citet{mimlre,lin2016neural,vashishth2018reside} adopted a multi-instance relation extraction setting, where multiple sentences are used to capture features associated with a pair of entities. These approaches assume that entities have already been identified and focus solely on classifying relations between entity pairs.

\citet{katiyar2016investigating,miwa2016end,bekoulis2018joint,dat2019end,nayak2020ptrnetdecoding} tried to bring the named entity recognition task and relation classification task together. \newcite{zheng2017joint} used a sequence tagging scheme to jointly extract the entities and relations. \citet{zeng2018copyre,nayak2020ptrnetdecoding} proposed an encoder-decoder model to extract relational triples with overlapping entities. \citet{hrlre2019takanobu} proposed a joint extraction model based on hierarchical reinforcement learning (HRL). 

With the introduction of pre-trained models such as BERT \cite{Devlin2019BERTPO}, many models used such models as sentence encoder to improve their performance. Models such as TPLinker \cite{Wang2020TPLinkerSJ}, CasRel \cite{Wei2020ANC}, TDEER \cite{Li2021TDEERAE}, PRGC \cite{Zheng2021PRGCPR}, PFN \cite{Yan2021APF}, GRTE \cite{Ren2021ANG}, OneRel \cite{Shang2022OneRelJE}, and BiRTE \cite{Ren2022ASB} use BERT\_{base} \cite{Devlin2019BERTPO} as an encoder and proposed table-filling method or relation specific tagging mechanism for joint entity and relation extraction. These models show remarkable performance on the NYT datasets in the restrictive experimental setting without considering the zero-cardinal sentences.

\section{Conclusion}

In this work, we present an innovative and challenging experiment design for relation extraction, which incorporates sentences containing zero triples (referred to as zero-cardinal sentences) in the dataset. We conduct comprehensive experiments involving 9 state-of-the-art (SOTA) models using the widely-used New York Times datasets. To tackle this task, we devise both an end-to-end modeling approach and a two-step modeling approach. 

During our investigations, we make a significant observation in the end-to-end modeling, where we notice a drop in the F1 score by approximately 10-15\% and 6-14\% in two versions of the NYT dataset. To address this issue, we propose the integration of a BERT-based classifier as an additional step for this task. Remarkably, this approach either achieves performance comparable to the end-to-end approach or even surpasses it.

We believe that our benchmark, focusing on relational triple extraction with zero-cardinality, will prove immensely valuable for future research in this domain. By introducing this unique experiment design, we aim to stimulate further advancements and foster progress in this field.

\section{Limitations}

One limitation of this work is that we benchmark this task using 9 SOTA joint models. There are many other SOTA models published in this area but it is difficult to benchmark all of them. We chose the 9 models in such as way that different kind of design choices in these models are represented in our study. We chose Seq2Seq model \cite{nayak2020ptrnetdecoding}, horn tagging-based models \cite{Wang2020TPLinkerSJ,Shang2022OneRelJE}, 0/1 tagging-based models \cite{Wei2020ANC,Li2021TDEERAE}, table-filling models \cite{Ren2021ANG} for rigorous study of this area.

\section{Ethics Statements}

Our work does not have any ethical concerns.

\bibliography{anthology,custom}
\bibliographystyle{acl_natbib}

\appendix
\section{Appendix}
\label{sec:appendix}

\subsection{GenBench Evaluation Cards}

\newcommand{\tabularwidth}{\textwidth}

\newcommand{\expone}{$\circ$}
        
\begin{table*}[hbt!]
\renewcommand{\arraystretch}{1.1}         
\setlength{\tabcolsep}{0mm}         
\begin{tabular}{|p{\tabularwidth}<{\centering}|}         
\hline
               
\rowcolor{gray!60}               
\textbf{Motivation} \\               
\footnotesize
\begin{tabular}{p{0.25\tabularwidth}<{\centering} p{0.25\tabularwidth}<{\centering} p{0.25\tabularwidth}<{\centering} p{0.25\tabularwidth}<{\centering}}                        
\textit{Practical} & \textit{Cognitive} & \textit{Intrinsic} & \textit{Fairness}\\
\expone\hspace{0.8mm}		
& 		
& 		
& 		

\vspace{2mm} \\
\end{tabular}\\
               
\rowcolor{gray!60}               
\textbf{Generalisation type} \\               
\footnotesize
\begin{tabular}{m{0.21\tabularwidth}<{\centering} m{0.2\tabularwidth}<{\centering} m{0.13\tabularwidth}<{\centering} m{0.13\tabularwidth}<{\centering} m{0.13\tabularwidth}<{\centering} m{0.2\tabularwidth}<{\centering}}                   
\textit{Compositional} & \textit{Structural} & \textit{Cross Task} & \textit{Cross Language} & \textit{Cross Domain} & \textit{Robustness}\\
& 		
& 		
& 		
& 		
& \expone\hspace{0.8mm}		

\vspace{2mm} \\
\end{tabular}\\
             
\rowcolor{gray!60}             
\textbf{Shift type} \\             
\footnotesize
\begin{tabular}{p{0.25\tabularwidth}<{\centering} p{0.25\tabularwidth}<{\centering} p{0.25\tabularwidth}<{\centering} p{0.25\tabularwidth}<{\centering}}                        
\textit{Covariate} & \textit{Label} & \textit{Full} & \textit{Assumed}\\  
& \expone\hspace{0.8mm}		
& 		
& 		

\vspace{2mm} \\
\end{tabular}\\
             
\rowcolor{gray!60}             
\textbf{Shift source} \\             
\footnotesize
\begin{tabular}{p{0.25\tabularwidth}<{\centering} p{0.25\tabularwidth}<{\centering} p{0.25\tabularwidth}<{\centering} p{0.25\tabularwidth}<{\centering}}                          
\textit{Naturally occuring} & \textit{Partitioned natural} & \textit{Generated shift} & \textit{Fully generated}\\
\expone\hspace{0.8mm}			
& 	
& 		
& 		

\vspace{2mm} \\
\end{tabular}\\
             
\rowcolor{gray!60}             
\textbf{Shift locus}\\             
\footnotesize
\begin{tabular}{p{0.25\tabularwidth}<{\centering} p{0.25\tabularwidth}<{\centering} p{0.25\tabularwidth}<{\centering} p{0.25\tabularwidth}<{\centering}}                         
\textit{Train--test} & \textit{Finetune train--test} & \textit{Pretrain--train} & \textit{Pretrain--test}\\
\expone\hspace{0.8mm}		
& 		
& 		
& 		

\vspace{2mm} \\
\end{tabular}\\

\hline
\end{tabular}
\label{tab:datacard}
\caption{We characterise all our experiments of Section 5 ($\circ$) in this datacard.}
\end{table*}

\end{document}